\pdfoutput=1  
\documentclass[journal,hideappendix]{vgtc}        

\onlineid{1175}

\vgtccategory{Area 5: Data Transformations}

\title{CycleChart: A Unified Consistency-Based Learning Framework for Bidirectional Chart Understanding and Generation}

\author{%
  Dazhen Deng,
  Sen Yang,
  Yuchen He,
  Yuan Tian, and
  Yingcai Wu
}

\authorfooter{
  \item
    Dazhen Deng, Sen Yang, Yuchen He, Yuan Tian, and Yingcai Wu are with
    Zhejiang University.
    E-mail: \{dengdazhen, 22451267, heyuchen, yuantian, ycwu\}@zju.edu.cn.
}

\abstract{%
  Current chart-related tasks, such as chart generation (NL2Chart), chart schema parsing, chart data parsing, and chart question answering (ChartQA), are typically studied in isolation, preventing models from learning the shared semantics that link chart creation and interpretation. We introduce CycleChart, a consistency-based learning framework for bidirectional chart understanding and generation. Unlike conventional multi-task approaches that draw training samples independently across tasks, CycleChart organizes all tasks around each single data instance. From a source table and natural-language query, the model generates a chart specification, renders and executes it, then learns to recover the schema and underlying data from the resulting chart image. This per-instance lifecycle design lets the model capture the full chain of transformations, from raw data through visual encoding to structured recovery, and a generate--parse consistency objective enforces semantic alignment between the forward generation and reverse parsing directions. To support this framework, we construct CycleChart-Bench, a lifecycle-aligned benchmark where every chart sample carries aligned annotations for generation, schema parsing, data parsing, and question answering. CycleChart achieves strong results across all four tasks and transfers effectively to unseen external benchmarks, demonstrating improved cross-task generalization and marking a step toward more general chart understanding models.
}


\teaser{
  \centering
  \includegraphics[width=\linewidth]{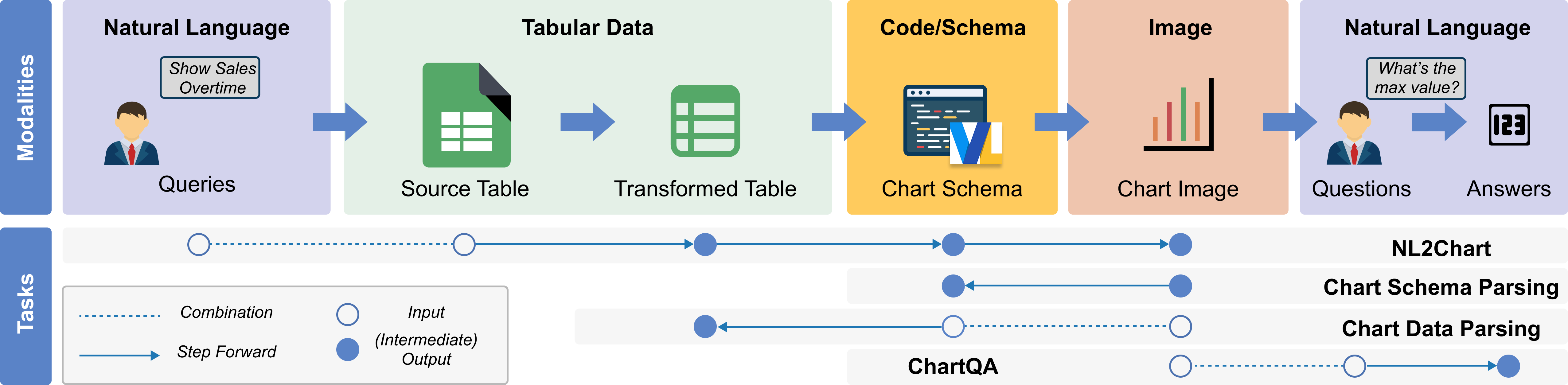}
  \caption{%
    The chart lifecycle spans four modalities (natural language, tabular data, code/schema, and image), forming a bidirectional pipeline that connects chart generation and chart understanding. NL2Chart traverses the forward path from a query and source table to a rendered chart; Chart Schema Parsing and Chart Data Parsing invert this path, recovering the specification and data from the image; ChartQA reasons over the chart to produce answers. CycleChart unifies all four tasks within a single model and enforces generate--parse consistency across the forward and reverse directions, bridging modalities into a closed cycle.%
  }
  \label{fig:tasks}
}

\graphicspath{{figs/}{figures/}{pictures/}{images/}{./}} 

\usepackage{makecell}
\usepackage{multirow}
\usepackage{tabularx}
\usepackage{array}
\usepackage{float}
\usepackage{booktabs}
\usepackage{amsmath}
\usepackage{amssymb}

\usepackage{mathptmx}                  

\newcommand{\impr}[1]{{\scriptsize\textcolor{gray}{#1}}}

\begin{document}


\firstsection{Introduction}

\maketitle

\label{sec:intro}
Charts encode numerical and categorical relationships through visual variables that together convey analytical semantics. Consequently, chart understanding and generation serve as the cornerstones of human-AI teaming for data analysis, as they determine whether models can not only interpret these visual artifacts and the embedded data insights, but also infer users' analytical intents.
Early chart understanding models, constrained by limited vision capabilities, relied heavily on OCR or component detection to provide explicit structural cues for reasoning. Similarly, chart generation has long been formulated as predicting chart parameters or code, without modeling how data semantics are transformed into visual encodings or why certain visual designs communicate specific analytical intents.

While recent advances in chart-related tasks~\cite{huang2024pixels} have demonstrated remarkable progress, these approaches still lack a unified semantic understanding of charts, as they typically model each task independently.
Chart Question Answering (ChartQA)~\cite{masry2022chartqa,masry2025chartqapro,wang2024charxiv,zheng2025advancing,vogel2025refchartqa} focuses on answering natural-language questions based on chart images, requiring both visual perception and numerical reasoning.
Chart Parsing~\cite{huang-etal-2024-lvlms,meng2024chartassisstant} aims to recover structured representations---such as chart schemas or underlying data tables---from rendered visualizations.
Chart Generation~\cite{tian2024chartgpt,luo2021nvbench,luo2025nvbench,narechania2020nl4dv,shuai2025deepvis} involves producing visual specifications or rendered charts from data tables or natural-language inputs.
To overcome such task fragmentation, the emergence of large multimodal models has motivated researchers to unify different chart-related tasks under a shared framework~\cite{xia2025chartx, han2023chartllama}.
However, these approaches mainly leverage the models' ability to handle multimodal inputs and outputs, rather than fully exploiting the intrinsic information in charts.
This raises a fundamental question: how can models move beyond multimodal alignment to genuinely capture the internal logic of charts?

From the perspective of visualization theory, charts can be viewed through the lens of the Grammar of Graphics (GoG~\cite{wilkinson2011grammar}), which defines a pipeline linking data and visual representations.
In this grammar, a data table is first transformed into analytical variables, which are then mapped to visual marks and channels to produce the final chart. The chart, in turn, conveys information that humans interpret into facts or insights (\autoref{fig:tasks}).
Each corresponds to a different stage along the data-visual-semantic pipeline: chart generation encodes data into visual forms, chart parsing decodes visuals back into structured representations, and chart question answering interprets visuals to extract semantic insights.
In practice, most datasets sample examples from one of these stages--either by generating charts from code or by collecting chart-question pairs from the web or academic publications--thus providing only partial supervision of the overall semantics.
As a result, current models learn fragmented mappings within the data-visual-semantic continuum, rather than understanding the complete reasoning process that connects data, visual encodings, and analytical interpretation.

This insight motivates a unified and self-consistent framework that leverages the supervision signals inherent in the chart creation process.
We propose \textbf{CycleChart}, a consistency-based learning framework that teaches MLLMs to perform both forward generation (from data and language to charts) and reverse parsing (from charts back to schema, data, and semantics) within a unified cycle.
Given natural language and data tables, the model first generates a chart schema and executes it through a deterministic renderer, producing a chart image and its visualization-level table.
These execution outputs then serve as automatically derived supervision targets: the model must recover the schema and data from the rendered chart, and the reconstruction loss enforces consistency between the generated and recovered representations.
Because the reverse-path targets are produced by executing the chart specification through a deterministic renderer rather than by manual annotation, this cycle yields automatically derived supervision that scales with the training data without additional labeling effort.

To support this framework, we introduce \textbf{CycleChart-Bench}, a benchmark that covers the \emph{full chart lifecycle}--from NL-driven chart generation to schema and data parsing, and finally to chart question answering.
CycleChart-Bench further includes both single-view and faceted multi-view charts, making generation and reasoning tasks substantially more challenging by requiring precise localization and comparison of values across multiple subplots. 
Evaluations on a wide range of open- and closed-source models show that these faceted, lifecycle-aligned samples expose significantly harder cases than prior benchmarks, demonstrating CycleChart-Bench as a rigorous and challenging testbed
for chart understanding. The contributions are as follows.
\begin{itemize}[leftmargin=*]
\item \textbf{CycleChart:} A consistency-based learning framework that unifies chart generation, schema parsing, data parsing, and question answering through a closed generate-parse cycle, where reverse-path supervision is automatically derived from renderer execution without manual annotation.
\item \textbf{CycleChart-Bench:} A lifecycle-aligned benchmark that links generation, parsing, and question answering within a cyclic loop, supporting both single-view and faceted charts.
\item \textbf{Insights:} Enforcing generate--parse consistency yields three findings:
(i)~per-instance alignment, not task diversity, drives the gains;
(ii)~the cycle objective benefits compositionally complex faceted charts most; and
(iii)~the framework transfers to unseen chart styles (CharXiv, nvBench VQL) without architecture changes.
\end{itemize}

\section{Related Work}
\label{related_work}
This section introduces the related studies on chart understanding, chart generation, and consistency-based learning.

\subsection{Chart Understanding}
Chart understanding bridges computer vision and natural language processing, with chart parsing as a major subtask. Chart parsing typically involves three components: component extraction, schema prediction, and data extraction. Component extraction detects axes, marks, labels, and text, usually via object detection and OCR. Schema prediction~\cite{poco2017reverse,chen2024onechart} aims to recover the chart's structural specification, expressed either as textual descriptions~\cite{tian2024chartgpt} or formal grammars such as matplotlib, Vega-Lite, or ggplot2. Data extraction~\cite{meng2024chartassisstant} is the most challenging step, requiring models to infer the precise values encoded by visual marks relative to chart axes.

Beyond structured parsing, another major research direction focuses on the textual interpretation of charts. Chart captioning~\cite{hsu2021scicap,tang2023vistext} generates natural-language summaries but suffers from ambiguity and loosely defined outputs. ChartQA addresses this limitation by answering questions about chart images through visual grounding and numerical reasoning~\cite{masry2022chartqa,masry2025chartqapro,wang2024charxiv,methani2020plotqa,kafle2018dvqa,xu2025chartpoint,xu2023chartbench,wu2024chartinsights,wei2025chartmind,masry2024chartinstruct,yang2025effective,vogel2025refchartqa}, covering tasks from simple chart-type identification to multi-step numerical reasoning over visual elements.

Recent unified frameworks aim to handle multiple chart-understanding tasks within a single model~\cite{zhang2024tinychart,masry2023unichart}. UniChart~\cite{masry2023unichart} introduces chart-specific pretraining tasks for both low-level element extraction and high-level reasoning, while ChartMoE~\cite{xu2024chartmoe} extends chart understanding to editing, replotting, highlighting, and transformation via natural-language instructions.
However, these methods largely focus on interpreting existing charts and do not complete the full semantic cycle connecting data, chart generation, and chart understanding. Consequently, the bidirectional link between data, visualization structure, and natural-language semantics remains underexplored.

\subsection{Chart Generation} Chart generation~\cite{ye2024generative} aims to predict chart schemas from data tables, either with or without natural-language descriptions.
When no textual intent is provided, the task becomes a chart recommendation, where the model analyzes a table to identify meaningful variable combinations and propose suitable visualizations~\cite{wongsuphasawat2015voyager,wongsuphasawat2017voyager,deng2023dashbot}.
With natural language guidance, recent work has increasingly focused on NL-driven chart generation~\cite{hoque2025natural}, progressing from syntactic parsing approaches~\cite{narechania2020nl4dv} to instruction-tuned LLMs trained on curated datasets and reward models~\cite{tian2024chartgpt,luo2021nvbench,luo2025nvbench,shuai2025deepvis}, enabling models to better translate analytic intents into chart specifications.

In addition, recent multimodal efforts have started to bridge chart understanding and chart generation within a single framework. ChartLlama~\cite{han2023chartllama} aligns chart, text, and data representations so that a multimodal LLM can perform comprehension tasks (e.g., QA, captioning, summarization) alongside generative tasks such as chart synthesis and editing.
Similarly, multi-task benchmarks like ChartVLM~\cite{xia2025chartx} evaluate perception, reasoning, and redrawing jointly, suggesting the value of integrated modeling. These works point toward integrated solutions but still leave open how to enforce stronger bidirectional consistency between generation and parsing---an issue we address with CycleChart.

\subsection{Consistency-based Learning}
A key idea in consistency-based learning is to enforce that applying a forward and reverse mapping sequentially reconstructs the input, encouraging invertible, structure-preserving transformations.
CycleGAN~\cite{zhu2017unpaired} introduced this idea for unpaired image translation, inspiring extensions such as DualGAN~\cite{yi2017dualgan}, ReCycleGAN~\cite{bansal2018recycle}, and dual learning in machine translation~\cite{he2016dual}.
The same principle has been applied to cross-modal tasks, where cycle-consistency aligns vision and language representations~\cite{cornia2018towards,wang2024cycle}.
These methods operate in continuous spaces and require a differentiable round-trip loss through the pixel or feature domain.
Charts, by contrast, are produced by executing a discrete specification through a deterministic renderer, making gradient-based cycle losses inapplicable and calling for a data-level consistency formulation.

In chart reasoning, grounding and consistency have been used to improve reliability.
RefChartQA~\cite{vogel2025refchartqa} introduces visual grounding for fine-grained chart QA, and Huang \textit{et al.}~\cite{huang-etal-2024-lvlms} align chart captions with underlying data to correct factual errors.
ChartPoint~\cite{xu2025chartpoint} further bridges visual perception and reasoning by integrating reflective interactions and re-rendered bounding boxes into chain-of-thought reasoning.
However, these studies remain one-directional---verifying text or visuals separately---without enforcing bidirectional semantic consistency between chart generation and understanding.
Our work explores this direction by applying generate--parse consistency at the data and specification level rather than in pixel space.

\section{Problem Formulation}

We consider four closely related tasks that together cover the full chart creation
and understanding lifecycle: NL2Chart, chart schema parsing, chart data parsing, and ChartQA.
We first introduce the shared notation and then define each task.
We distinguish two types of tables throughout:
the \emph{raw table} $t^{\text{raw}}$ is the original source data, while the \emph{visualization-level table} $t^{\text{vis}}$ is the result of applying data transformations (e.g., aggregation, filtering, binning) specified by the chart schema, which contains exactly the values encoded by the visual marks in the rendered chart.

Let $I$ denote the space of chart images (e.g., rasterized plots),
$T$ the space of structured data tables,
$Q$ the space of natural-language queries or instructions,
$S$ the space of chart schemas (e.g., Vega-Lite specifications),
and $A$ the space of answers.
We assume a deterministic rendering pipeline that takes a schema and the raw
table, internally applies the data transformations specified by the schema to
produce a visualization-level table, and render the final chart image:
\[
    R : S \times T \rightarrow I \times T,
    \quad
    (\hat{\imath},\; t^{\text{vis}}) = R(s,\; t^{\text{raw}}).
\]
Here $t^{\text{vis}}$ is a byproduct of the rendering execution, not a separate
input.  In our Vega-Lite instantiation, we extract $t^{\text{vis}}$ by
intercepting the transformed data from the Altair runtime.

\paragraph{Natural Language to Chart Generation (NL2Chart).}
Given a raw table $t^{\text{raw}}$ and a natural-language query $q$ describing the desired analytical intent, the model generates a chart schema $\hat{s}$:
\begin{equation}
    f_{\text{NL2Chart}} :
    T \times Q \rightarrow S,
    \quad
    \hat{s} = f_{\text{NL2Chart}}(t^{\text{raw}}, q).
\end{equation}
In the Vega-Lite instantiation, $\hat{s}$ is a declarative specification that
includes built-in data transformations (e.g., aggregation, filtering).
Executing $\hat{s}$ through the renderer yields the chart image and its
visualization-level table:
$(\hat{\imath},\; \hat{t}^{\text{vis}}) = R(\hat{s},\; t^{\text{raw}})$.

\paragraph{Chart Schema Parsing.}
Chart schema parsing aims to recover the schema from a rendered chart image.
In practice, the model also receives the table schema (column names and types) as auxiliary context, simulating a realistic scenario in which a user possesses the source dataset and wants to recover \emph{how} it was visualized:
\begin{equation}
    f_{\text{schema}} :
    I \times T_{\text{meta}} \rightarrow S,
    \quad
    \tilde{s} = f_{\text{schema}}(\hat{\imath},\; t_{\text{meta}}).
\end{equation}
The goal is for the predicted schema $\tilde{s}$ to match the visual schema
$s^{\text{vis}}$ (the gold specification with non-visual transform operations
removed) up to permissible structural variations.

\paragraph{Chart Data Parsing.}
Chart data parsing further extracts the visualization-level table from the chart.
Since the table depends on the schema's encodings (e.g., mark type, axes, facets),
the task takes both the image and schema as input:
\begin{equation}
    f_{\text{data}} :
    I \times S \rightarrow T,
    \quad
    \tilde{t}^{\text{vis}} = f_{\text{data}}(\hat{\imath}, \hat{s}).
\end{equation}
The supervision target is the visualization-level table
$t^{\text{vis}}$ extracted from the Vega-Lite rendering pipeline.

\paragraph{Chart Question Answering (ChartQA).}
Given a chart image $\hat{\imath}$ and a natural-language question $q$ about the chart, the model predicts a textual answer:
\begin{equation}
    f_{\text{QA}} :
    I \times Q \rightarrow A,
    \quad
    \hat{a} = f_{\text{QA}}(\hat{\imath}, q_{\text{QA}}).
\end{equation}
ChartQA requires both visual grounding and numerical or relational reasoning
over the content encoded in the chart.

\paragraph{From tasks to cycle.}
The four tasks above are not independent: the output of forward generation
($\hat{s}$) is executed by the renderer to produce the inputs for all three
reverse tasks ($\hat{\imath}$, $t^{\text{vis}}$), and the reverse tasks in
turn verify whether the generated chart faithfully encodes the intended
structure and semantics.
This bidirectional dependency forms a closed generate--parse cycle that we
exploit as a consistency-based training objective in
\autoref{sec:training}.

\section{CycleChart-Bench Construction}
\label{sec:benchmark}

To support our cycle-consistent training framework, we build \emph{CycleChart-Bench}, a multi-task benchmark that provides aligned supervision for NL2Chart, schema parsing, data parsing, and ChartQA.

\begin{figure}[tb]
    \centering
    \includegraphics[width=\linewidth]{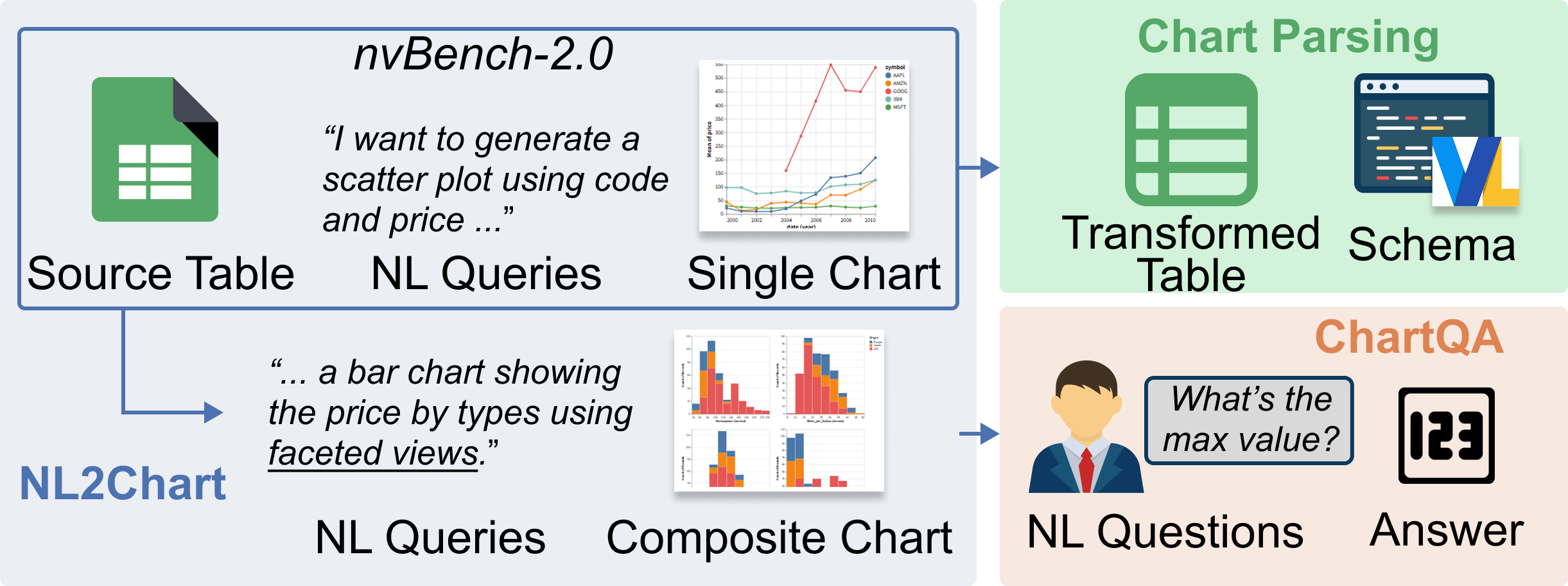}
    \caption{CycleChart-Bench construction. nvBench~2.0 queries produce single-view charts, while our facet-oriented queries generate composite charts. Both chart types supply transformed tables and QA pairs for aligned parsing and reasoning supervision.}
    \label{fig:benchmark}
\end{figure}

\paragraph{Motivation.}
Existing multi-task chart datasets often assemble tasks from heterogeneous sources, leading to weak coupling between a chart image, its underlying table, generation specification, and QA annotations.
Such partial supervision prevents models from learning how chart creation, structure recovery, and reasoning relate to one another.

CycleChart-Bench instead models the \emph{entire} lifecycle.
For each sample, we begin with a raw table and an NL query that specifies an analytic intent, derive a chart specification, and execute it through the Vega-Lite renderer to produce the chart image and its visualization-level table.
Parsing and QA labels are then derived automatically from these execution outputs: the specification serves as the schema parsing target, the visualization-level table as the data parsing target, and QA pairs are generated programmatically from the table values.
This construction requires no manual annotation of parsing or QA labels, ensuring tight cross-task alignment while remaining fully scalable.

Beyond providing automatically derived targets, the benchmark serves two roles that an online cycle alone cannot fulfill.
First, it supplies verified gold specifications that guarantee every training sample can complete the full generate--parse loop.
Second, pre-computing the execution outputs offline avoids invoking the Vega-Lite renderer at every training iteration, making large-scale cycle-consistent training practical.

\paragraph{Based on nvBench~2.0.}
Our benchmark is constructed on nvBench~2.0~\cite{luo2025nvbench}, which provides raw tables, NL queries, and Vega-Lite specifications for NL2Chart.
nvBench's diverse chart types and intent-driven queries make it a strong base, but it lacks parsing labels and QA supervision.
CycleChart-Bench extends each example into a full cycle:
\[
(t^{\text{raw}}, q) \rightarrow (t^{\text{vis}}, s, \hat{\imath})
\rightarrow \text{parsing labels}
\rightarrow \text{QA pairs}.
\]

\paragraph{Facet-Augmented Charts.}
To reflect real-world multi-view layouts, we augment the nvBench charts by creating faceted versions when suitable categorical fields are present.
These composed charts share data columns across subplots and enable cross-view reasoning, increasing structural diversity beyond single-view plots.

\paragraph{Visualization-Level Tables.}
For each (possibly faceted) Vega-Lite specification, we obtain the visualization-level table $t^{\text{vis}}$ by executing the spec with Altair and intercepting the transformed data used for mark rendering.
This yields precise supervision for chart data parsing.

\paragraph{ChartQA Generation.}
QA annotations are produced programmatically from $t^{\text{vis}}$, covering both
single-view reasoning (extrema, comparisons, simple aggregations) and cross-facet
reasoning for composed charts.
LLMs are used only to generate natural-language templates; all answers are computed
deterministically to avoid hallucination.

\paragraph{Statistics and validation.}
CycleChart-Bench contains 6,507 charts in total:
5,372 single-view and 1,135 faceted charts with five common mark types.
We use an 8:1:1 train/validation/test split
(5,205 / 650 / 652 samples), preserving the ratio of single- and multi-view charts.
All Vega-Lite specifications are compiled via \texttt{vl-convert} to verify
renderability; any specification that fails compilation is discarded.
QA answers are computed deterministically from $t^{\text{vis}}$, avoiding hallucinated labels.
During evaluation, invalid predicted schemas receive a zero score for both validity
and visual similarity metrics.

\subsection{Construction Details}

\paragraph{Chart type expansion and faceted chart augmentation.}
We first filter out unsupported or non-standard visualization types in nvBench~2.0, including boxplots and malformed specifications that fail to compile or do not conform to Vega-Lite conventions.
We then programmatically augment eligible specifications into faceted charts. For each original Vega-Lite spec, we identify categorical fields that can partition the visualization and inject corresponding facet encodings (e.g., \texttt{``column''} or \texttt{``row''}) directly into the \texttt{encoding} block.
All augmented specifications are rendered with \texttt{vl-convert} and validated through compile-time and runtime correctness checks.
This procedure yields 1,135 high-quality faceted charts, which substantially increase structural diversity and introduce multi-view reasoning challenges.

\paragraph{Parsing labels.}
For schema parsing, we remove all \texttt{transform} operations from the original specification.
These transformations (e.g., grouping, aggregation) modify the intermediate data but are not visually observable and therefore cannot be inferred from the image alone.
The resulting schema $s^{\text{vis}}$ represents the minimal visual structure needed for accurate parsing supervision.
For data parsing, we execute the specification's transformation pipeline via \texttt{pandas} and serialize the result as CSV, capturing exactly the values encoded by the chart marks.

\paragraph{ChartQA annotations.}
We generate QA pairs using a large VLM (Qwen3-VL-235B). The model is provided with (i)~the rendered image, (ii)~the full Vega-Lite schema, (iii)~the ground-truth data table produced by our parsing pipeline, and (iv)~several seed questions.
It is then instructed to produce new question--answer pairs that follow the same reasoning patterns while remaining faithful to the chart content.
Seed questions are retrieved using a RAG pipeline: we embed 1,000 reasoning questions from CharXiv using \texttt{intfloat/multilingual-e5-large} and build a vector store.
For each chart, we query the store using its NL intent to obtain the top-3 semantically relevant questions as stylistic exemplars.
All final answers are verified deterministically against $t^{\text{vis}}$: we recompute each answer from the visualization-level table and discard any QA pair whose VLM-generated answer deviates from the recomputed value.
This procedure retains approximately 78\% of the generated pairs, ensuring that every retained answer is grounded in the chart data rather than hallucinated by the VLM.

\begin{figure*}[!htbp]
    \centering
    \includegraphics[width=\linewidth]{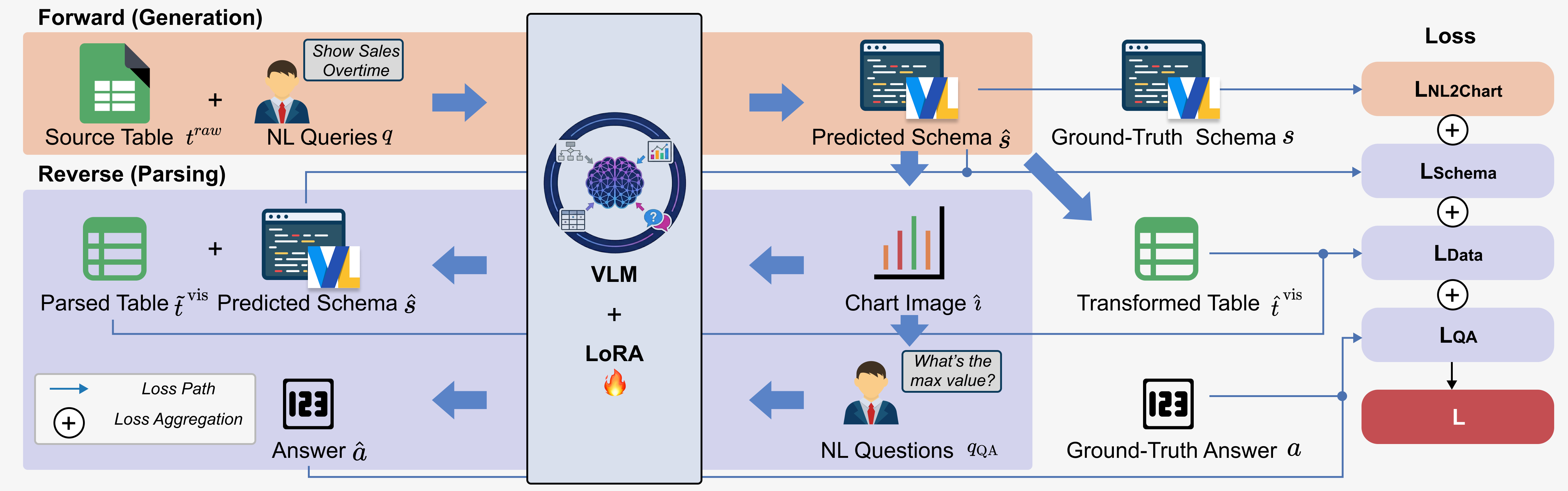}
    \caption{Overview of the CycleChart training framework. Given a source table and an NL query, the model generates a chart specification (NL2Chart), which is executed to obtain a chart image and its visualization-level table. The model then parses the generated chart into a schema and table (Chart Parsing) and answers NL questions about the chart (ChartQA). All tasks use cross-entropy supervision, forming a closed generate--parse cycle that enforces consistency.}
    \label{fig:model}
\end{figure*}

\section{CycleChart Training Framework}
\label{sec:training}

CycleChart trains a single multimodal model to jointly perform chart generation and chart understanding within a closed generate--parse cycle.
Each training iteration operates on a single data instance and traverses both directions of this cycle, as illustrated in \autoref{fig:model}: the \emph{forward} path generates a chart from data and language, while the \emph{reverse} path recovers the specification, data, and semantics from the rendered chart image.
Crucially, the supervision targets for the reverse path are not manually annotated: they are automatically derived by executing the chart specification through a deterministic renderer, eliminating the need for additional labeling.

\paragraph{Forward: Chart Generation (NL2Chart).}
Given a raw table $t^{\text{raw}}$ and a natural-language query $q$, the model predicts a chart schema $\hat{s}$, supervised by a token-level cross-entropy loss against the ground-truth specification:
\[
L_{\text{NL2Chart}} = \mathrm{CE}(\hat{s}, s).
\]
We additionally validate whether $\hat{s}$ is syntactically correct and renderable; only valid schemas proceed to the reverse path, because an unrenderable specification cannot produce the chart image and visualization-level table on which the consistency cycle depends.
For valid schemas, the Vega-Lite renderer $R$ executes $\hat{s}$ on the raw table $t^{\text{raw}}$, producing a chart image and its visualization-level table as execution byproducts:
$(\hat{\imath},\; \hat{t}^{\text{vis}}) = R(\hat{s},\; t^{\text{raw}})$.
Both $\hat{\imath}$ and $\hat{t}^{\text{vis}}$ then serve as the input and supervision targets for the reverse path.
In practice, CycleChart-Bench pre-computes these execution outputs offline, avoiding renderer invocations during training and enabling efficient cycle-consistent learning at scale (see \autoref{sec:benchmark}).

\paragraph{Reverse: Chart Parsing and Reasoning.}
Starting from the rendered chart image $\hat{\imath}$, the model performs three tasks that invert or interpret the forward generation.
The reverse-path targets, which are the visual schema $s^{\text{vis}}$ (the gold specification with non-visual \texttt{transform} operations removed; see \autoref{sec:benchmark}) and the visualization-level table $t^{\text{vis}}$, are both renderer-derived execution byproducts, as described in the opening paragraph.

\emph{Chart Schema Parsing.}
The model recovers a schema $\tilde{s}$ from $\hat{\imath}$.
The target $s^{\text{vis}}$ retains only the visually observable structure of the gold specification, so the reconstruction loss enforces generate--parse consistency at the level of visual encoding:
\[
L_{\text{Schema}} = \mathrm{CE}(\tilde{s},\; s^{\text{vis}}).
\]

\emph{Chart Data Parsing.}
Conditioned on the rendered image and the visual schema, the model predicts a visualization-level table $\tilde{t}^{\text{vis}}$.
The target $t^{\text{vis}}$ is extracted from the Vega-Lite execution pipeline:
\[
L_{\text{Data}} = \mathrm{CE}(\tilde{t}^{\text{vis}},\; t^{\text{vis}}).
\]
Together with $L_{\text{Schema}}$, this reconstruction loss completes the full generate--parse cycle, requiring the model to faithfully recover both the visual structure and the underlying data from the rendered chart.
During training, the gold visual schema $s^{\text{vis}}$ is provided as input to data parsing; at test time, the model uses its own predicted schema instead.
This gap is mitigated in practice because schema parsing already achieves high accuracy (see \autoref{tab:CycleChart_schema}), limiting error propagation.

\emph{ChartQA.}
Given the chart image and a QA query $q_{\text{QA}}$, the model predicts an answer $\hat{a}$, supervised by:
\[
L_{\text{QA}} = \mathrm{CE}(\hat{a}, a).
\]
Schema and Data Parsing verify that the model can faithfully recover the \emph{structural} content of a chart, i.e., its visual encoding and underlying values.
ChartQA complements them by testing \emph{semantic} consistency: whether the model can derive correct analytical conclusions from the same visual artifact.
Together, the three reverse tasks form a progression from structure to semantics, ensuring that the generate--parse cycle enforces consistency at every level of chart interpretation.
QA targets are generated programmatically and verified against $t^{\text{vis}}$ (see \autoref{sec:benchmark}), so the supervision remains grounded in the renderer's execution outputs.

\paragraph{Overall Objective.}
The full set of per-instance losses spans forward generation and reverse parsing:
\begin{equation}
    \mathcal{L}
    =
    \underbrace{L_{\text{NL2Chart}}}_{\text{forward (generate)}}
    +
    \underbrace{L_{\text{Schema}}
    +
    L_{\text{Data}}
    +
    L_{\text{QA}}}_{\text{reverse (parse)}} .
\end{equation}
All four losses use token-level cross-entropy without additional weighting.
In each iteration, a single task is sampled, 80\% from the three consistency tasks (NL2Chart, Schema Parsing, Data Parsing) and 20\% from ChartQA. Only the corresponding loss term is computed.

The key design principle is \emph{per-instance alignment}: all tasks for a given data instance share the same chart image and specification, so the forward generation and reverse parsing are always grounded in the same visual artifact.
This is what distinguishes CycleChart from standard multi-task fine-tuning, where samples for different tasks are drawn independently and this alignment is lost.
The consistency operates at the \emph{data level}. The deterministic renderer bridges the forward and reverse paths by producing exact supervision targets from the generated specification, rather than requiring a differentiable round-trip loss as in classical cycle-consistency formulations for continuous domains.
This simplicity is a feature: because every reverse-path target is an exact execution output with no annotation gap, the model receives a noise-free consistency signal that scales trivially with data, requires no reward modeling, and converges in as few as 400 steps (\autoref{fig:steps}).

\section{Experiments}

We evaluate CycleChart as a unified framework for chart generation, parsing, and reasoning.
We conduct evaluations on CycleChart-Bench and several external benchmarks, compare against strong VLMs and chart-specialized models, and perform ablations across different backbones and consistency schedules.

\subsection{Implementation Details}

We implement CycleChart on top of an instruction-tuned vision-language backbone.
All models are trained with the same tokenizer and visual encoder as the underlying backbone.
During training, each sample triggers one forward pass that computes the losses for the corresponding task (NL2Chart, Schema Parsing, Data Parsing, or ChartQA), depending on the sampling strategy described in \autoref{sec:training}.

\paragraph{Vega-Lite rendering and data extraction.}
We adopt Altair as a front-end to the Vega-Lite compiler.
For every predicted or ground-truth specification, we validate syntactic correctness and, if renderable, compile it into a visualization.
Visualization-level tables $t^{\text{vis}}$ are extracted directly from the Vega-Lite runtime, ensuring that data parsing is supervised with the exact values encoded in the chart marks.

\paragraph{Training setup.}
We fine-tune our models using AdamW with a constant learning rate of $1\times10^{-5}$ and a per-device batch size of~2.
LoRA-based parameter-efficient tuning (rank\,=\,16, $\alpha$\,=\,32, dropout\,=\,0.05) is applied to both the \textit{language model layers} and the \textit{vision--language projector}, while the vision encoder and all other components remain frozen.
We train for 2,000 steps with a constant learning-rate scheduler, allocating approximately 80\% of the training steps to
consistency-oriented tasks (NL2Chart, Schema Parsing, Data Parsing) and 20\% to ChartQA.
Training is performed on a single A100 GPU and we keep the total number of training steps constant in all ablations.
All reported results are from a single training run; the low variance across our five-backbone ablation (\autoref{tab:abl_xbench_backbones}), where every backbone improves consistently, suggests that the findings are robust.

\paragraph{Inference.}
We adopt a deterministic decoding setup for all evaluations to ensure fair and reproducible comparisons.
Unless otherwise specified, inference uses greedy decoding (temperature\,=\,0, top-$k$\,=\,1, top-$p$\,=\,1.0) with a maximum
output length of 2{,}048 tokens.

\newcolumntype{L}{>{\raggedright\arraybackslash}p{3.2cm}}
\newcolumntype{V}{>{\centering\arraybackslash}p{2.4cm}}
\newcolumntype{C}{>{\centering\arraybackslash}p{1.8cm}}
\newcolumntype{Q}{>{\raggedright\arraybackslash}X}

\begin{table*}[htb]
\centering
\small
\caption{Zero-shot performance on NL2Chart tasks. External models are directly evaluated without fine-tuning. Bold indicates the best performance, and underline indicates the second best.}
\label{tab:CycleChart_nl2chart}
\begin{tabularx}{\textwidth}{XVVVVV}
\toprule
\multirow{2}{*}{\textbf{Model}} &
\multicolumn{5}{c}{\textbf{NL2Chart}}\\
\cmidrule(lr){2-6}
& ROUGE-L recall $\uparrow$ & PSNR $\uparrow$ & CLIP $\uparrow$ & MS-SSIM $\uparrow$ & Valid $\uparrow$ \\
\midrule
\multicolumn{6}{l}{\textit{General VLMs}} \\
claude-3.5-sonnet     & 0.7666 & 20.8410 & 0.6821 & 0.4495 & 79.60\% \\
gemini-2.5-flash      & 0.7987 & 10.6326 & 0.6886 & 0.4282 & 84.51\% \\
gemini-2.5-pro        & 0.7752 & 21.1006 & 0.7678 & 0.5097 & 89.42\% \\
glm-4.5v              & 0.6894 & 10.3034 & 0.4727 & 0.2826 & 54.75\% \\
gpt-4.1               & 0.7369 & 17.2544 & 0.6909 & 0.4576 & 80.83\% \\
\midrule
\multicolumn{6}{l}{\textit{Our Models (Trained on CycleChart-Bench)}} \\
CycleChart-3B         & \underline{0.8062} & \underline{35.4636} & \underline{0.8739} & \underline{0.6180} & \underline{96.93\%} \\
\textbf{CycleChart-7B} &
\textbf{0.8712} & \textbf{43.8004} & \textbf{0.9098} & \textbf{0.6759} & \textbf{98.93\%}\\
\bottomrule
\end{tabularx}
\end{table*}

\subsection{Main Results}
We use CycleChart-Bench as a unified benchmark to evaluate a wide range of models---including general-purpose VLMs and chart-specific models---and compare their performance against our CycleChart models. 
The goal is to assess how well existing models handle chart generation, parsing, and reasoning, and to quantify the advantage brought by our consistency-based training.

\paragraph{Baseline Selection. }
We select a diverse set of baseline models for comparison, including both general-purpose VLMs and chart-specialized models. The general-purpose VLMs include advanced commercial proprietary models like Gemini-2.5-Flash~\cite{comanici2025gemini}, Gemini-2.5-Pro~\cite{comanici2025gemini}, Claude-3.5-Sonnet~\cite{claude3_5}, GLM-4.5v~\cite{vteam2025glm45v}, and GPT-4.1~\cite{gpt4_1}, which are known for their strong multimodal capabilities. The chart-specialized models include ChartInstruct-LLaMA2~\cite{masry2024chartinstruct}, TinyChart-3B-768~\cite{zhang2024tinychart}, and unichart-chartqa-960~\cite{masry2023unichart}, which are specifically designed for chart-related tasks.

\begin{table*}[htbp]
\centering
\small
\caption{Zero-shot performance on Chart Parsing and ChartQA tasks. Bold and underline indicate the best and second-best performance.}
\label{tab:CycleChart_schema}
\setlength{\tabcolsep}{4pt}
\begin{tabularx}{\textwidth}{XVCCCCVC}
\toprule
\multirow{2}{*}{\raisebox{-2pt}{\textbf{Model}}} &
\multicolumn{5}{c}{\textbf{Chart Schema Parsing}} &
\textbf{Chart Data Parsing} &
\textbf{ChartQA Task} \\
\cmidrule(lr){2-6}
& ROUGE-L recall $\uparrow$ & PSNR $\uparrow$ & CLIP $\uparrow$ & MS-SSIM $\uparrow$ & Valid $\uparrow$ &RNSS $\uparrow$&EM $\uparrow$\\
\midrule
\multicolumn{7}{l}{\textit{General VLMs}} \\
claude-3.5-sonnet     & 0.8894 & 25.1227 & 0.7994 & 0.6030 & 84.82\% & 69.9386 & 0.5271 \\
gemini-2.5-flash      & \underline{0.9509} & 33.7034 & 0.8844 & 0.6886 & 91.41\% & 72.5527 & 0.6713\\
gemini-2.5-pro        & 0.9455 & 54.5638 & 0.8975 & 0.7846 & 91.41\% & 70.8386 & 0.6713\\
glm-4.5v              & 0.8704 & 20.8042 & 0.5321 & 0.3970 & 56.29\% & 70.6999 & 0.6015\\
gpt-4.1               & 0.9291 & 23.1719 & 0.6871 & 0.5128 & 71.63\% & 75.5297 & \underline{0.6744}\\
\midrule
\multicolumn{7}{l}{\textit{Chart-Specialized Models}} \\
ChartInstruct-Llama2  & -- & -- & -- & -- & -- & -- & 0.3968\\
TinyChart-3B-768      & -- & -- & -- & -- & -- & 46.5612& 0.3131\\
\midrule
\multicolumn{7}{l}{\textit{Our Models (Trained on CycleChart-Bench)}} \\
CycleChart-3B         & 0.9443 & \underline{77.2901} & \underline{0.9813} & \underline{0.9011} & \underline{99.85\%} & \underline{90.8364} & 0.6589\\
\textbf{CycleChart-7B} &
\textbf{0.9732} & \textbf{89.2574} & \textbf{0.9973} & \textbf{0.9745} & \textbf{100.00\%} & \textbf{95.2947} & \textbf{0.7550}\\
\bottomrule
\end{tabularx}
\end{table*}

\paragraph{Metrics.} We adopt task-specific metrics:
\begin{itemize}[leftmargin=*]
    \item \textbf{NL2Chart \& Chart Schema Parsing.}
    Both produce a Vega-Lite specification, evaluated from three perspectives.
    \emph{Textual similarity}: we compute ROUGE-L recall~\cite{lin2004rouge} on normalized (key-sorted) JSON strings to measure whether the generated specification covers the essential components of the ground truth.
    \emph{Visual similarity}: we render both specifications as same-size images and compute PSNR~\cite{5596999} (pixel-level fidelity), MS-SSIM~\cite{1292216} (multi-scale structural consistency), and CLIP~\cite{radford2021learningtransferablevisualmodels} (perceptual and semantic alignment); unrenderable specifications receive a zero score.
    \emph{Validity}: the percentage of specifications that can be successfully rendered without errors.
    \item \textbf{Chart Data Parsing.}
    We use Relative Number Set Similarity (RNSS), following ChartQA~\cite{masry2022chartqa}, which measures the overlap between predicted and ground-truth data values while accounting for numerical scale.
    Although Relative Mapping Similarity (RMS)~\cite{zhang2024tinychart} is a stronger metric, our extracted tables cannot always be parsed as key-value mappings, so we adopt RNSS for generality.
    RNSS intentionally penalizes structural mismatches (e.g., wide vs.\ long format), which is appropriate because our Data Parsing task is \emph{schema-conditioned}: deviations indicate a failure to follow the visual specification rather than a harmless format difference.
    \item \textbf{ChartQA task.}
    We use tolerance-aware Exact Match (EM): numeric predictions are accepted within a small absolute or relative threshold, while textual answers are matched after normalization.
\end{itemize}

\paragraph{Analysis.}
For NL2Chart (\autoref{tab:CycleChart_nl2chart}), CycleChart-7B achieves the best performance across all textual, visual, and validity metrics.
The 3B variant performs strongly, showing that our consistency-based training reliably produces both textually correct and visually faithful specifications.
Among general VLMs, Gemini-2.5-pro offers the strongest visual similarity, while Claude-3.5-sonnet, Gemini-2.5-flash, and GPT-4.1 achieve reasonable textual overlap and validity.
Chart-specific open-source models (e.g., ChartInstruct, UniChart, TinyChart) fail to produce valid Vega-Lite specifications; as they are not designed for forward chart generation, we exclude them from NL2Chart reporting.

For chart schema parsing (\autoref{tab:CycleChart_schema}), CycleChart-7B again achieves the best performance, with the 3B version obtaining a strong result, highlighting the benefits of our training pipeline.
Apart from CycleChart, general VLMs such as Gemini-2.5-pro and Gemini-2.5-flash also perform reasonably well on this task.
ChartInstruct-LLaMA2, UniChart-chartqa-960, and TinyChart once again fail to follow the parsing instructions and cannot produce valid schemas.

For chart data parsing (\autoref{tab:CycleChart_schema}), we follow the pipeline described in \autoref{sec:training}: at test time the model first predicts a schema from the chart image, then extracts the data table conditioned on its own predicted schema (not the gold one).
CycleChart-7B achieves an RNSS score of 95.29, significantly outperforming all existing models. The 3B variant also attains a strong score of 90.84, demonstrating the effectiveness of our consistency-based training in enhancing data extraction capabilities.
Given larger model sizes and higher training data quality, proprietary VLMs like Gemini-2.5-pro and Claude-3.5-sonnet perform better than open-source chart-specific models. ChartInstruct-LLaMA2 and unichart-chartqa-960 are unable to follow the parsing instructions effectively and thus fail to produce valid outputs for this task.

For the ChartQA task (\autoref{tab:CycleChart_schema}), CycleChart-7B reaches 0.7550 EM, beyond the performance of proprietary VLMs such as GPT-4.1 and Gemini-2.5-Pro.
The 3B variant achieves 0.6589 EM, slightly below GPT-4.1 and Gemini-2.5-Pro but clearly outperforming all open-source chart-specialized baselines.
These results show that consistency-based training substantially enhances chart reasoning and enables CycleChart to achieve strong zero-shot performance across diverse chart question-answering scenarios.

\subsection{Ablation Study}
\begin{table*}[h]
\centering
\small
\setlength{\tabcolsep}{4pt}
\begin{tabular}{l cc cc cc cc cc}
\toprule
\textbf{Tasks}
& \multicolumn{2}{c}{\textbf{NL2Chart}}
& \multicolumn{2}{c}{\textbf{Chart Data Parsing}}
& \multicolumn{6}{c}{\textbf{ChartQA Task}} \\
\cmidrule(lr){2-3}\cmidrule(lr){4-5}\cmidrule(lr){6-11}
\textbf{Backbone}
& \makecell{NVBench\\[-2pt]{\scriptsize Rouge-L$\uparrow$}} & \makecell{+Consist.\\[-2pt]{\scriptsize ($\Delta$)}}
& \makecell{ChartMOE-Align\\[-2pt]{\scriptsize Relax Acc.$\uparrow$}} & \makecell{+Consist.\\[-2pt]{\scriptsize ($\Delta$)}}
& \makecell{ChartQA\\[-2pt]{\scriptsize EM$\uparrow$}} & \makecell{+Consist.\\[-2pt]{\scriptsize ($\Delta$)}}
& \makecell{ChartQA-Pro\\[-2pt]{\scriptsize EM$\uparrow$}} & \makecell{+Consist.\\[-2pt]{\scriptsize ($\Delta$)}}
& \makecell{CharXiv\\[-2pt]{\scriptsize EM$\uparrow$}} & \makecell{+Consist.\\[-2pt]{\scriptsize ($\Delta$)}} \\
\midrule
LLaVA-OV-4B
& .683 & .691 \impr{(+.008)}
& .746 & .748 \impr{(+.003)}
& .842 & .843 \impr{(+.001)}
& .365 & .371 \impr{(+.006)}
& .418 & .436 \impr{(+.018)} \\
Qwen2.5-VL-3B
& .705 & .717 \impr{(+.012)}
& .682 & .714 \impr{(+.032)}
& .780 & .804 \impr{(+.024)}
& .295 & .319 \impr{(+.024)}
& .349 & .358 \impr{(+.009)} \\
Qwen2.5-VL-7B
& .681 & .717 \impr{(+.036)}
& .729 & \textbf{.758} \impr{(+.030)}
& .800 & \textbf{.846} \impr{(+.046)}
& .339 & \textbf{.377} \impr{(+.039)}
& .371 & .429 \impr{(+.058)} \\
Qwen3-VL-4B
& .701 & .709 \impr{(+.008)}
& .718 & .751 \impr{(+.033)}
& .774 & .811 \impr{(+.037)}
& .307 & .340 \impr{(+.033)}
& .428 & .444 \impr{(+.016)} \\
Qwen3-VL-8B
& .711 & \textbf{.723} \impr{(+.012)}
& .751 & .756 \impr{(+.005)}
& .725 & .824 \impr{(+.098)}
& .296 & .348 \impr{(+.052)}
& .397 & \textbf{.472} \impr{(+.075)} \\
\bottomrule
\end{tabular}
\caption{\textbf{Ablation across backbones on external benchmarks (zero-shot).}
Each backbone is evaluated as-is and with our generate--parse \emph{consistency} training (+Consist.), trained only on \textbf{CycleChart-Bench} with no tuning on the target benchmark.
Parenthesized values denote absolute improvement.}
\label{tab:abl_xbench_backbones}

\end{table*}

We further evaluate the generalizability of our generate--parse consistency objective beyond CycleChart-Bench.
To avoid evaluating only on the dataset used for training, we conduct ablation studies exclusively on \emph{external} benchmarks spanning chart QA, parsing, and NL2Chart tasks.  
This setup enables us to isolate the effect of consistency-based training and verify that the performance gains are not tied to dataset-specific patterns but reflect real improvements in multi-task chart understanding and reasoning.
The results are summarized in \autoref{tab:abl_xbench_backbones}.

\paragraph{Baseline Selection.}
Our main experiments use Qwen2.5-VL-7B-Instruct as the default backbone; accordingly, Qwen2.5-VL-3B/7B-Instruct serve as the primary baselines in this ablation.
To further test generality, we additionally include LLaVA-OneVision-1.5-4B-Instruct (a different architecture) and the more recent Qwen3-VL-4B/8B-Instruct, which were released after our framework was developed.
Together, the five backbones span two model families, three architectures, and capacities from 3B to 8B.

\paragraph{External Datasets and Metrics.}
Our evaluation spans three major task categories:
\begin{itemize}[leftmargin=*,nosep]
    \item \textbf{NL2Chart:} evaluated with \textbf{Rouge-L} on nvBench~\cite{luo2021nvbench}, measuring overlap between generated and reference VQL outputs.
    \item \textbf{Chart Data Parsing:} evaluated on chart-to-table examples from ChartMOE-Align~\cite{xu2024chartmoe} with \textbf{Relax Acc.}, measuring key alignment and cell-level correctness under mild numeric/textual tolerances.
    \item \textbf{ChartQA:} evaluated on ChartQA~\cite{masry2022chartqa}, ChartQA-Pro~\cite{masry2025chartqapro}, and CharXiv~\cite{wang2024charxiv} (QA only; schema parsing is not evaluated on CharXiv, as its charts are not Vega-Lite) with \textbf{Exact Match (EM)}, combining numeric and textual matching with minor tolerance.
\end{itemize}

\paragraph{Analysis.}
Consistency training yields improvements on every benchmark across all five backbones, confirming the generality of the generate--parse objective.

\textit{Cross-generation comparison at a similar scale.}
Comparing Qwen2.5-VL and Qwen3-VL at similar model sizes reveals complementary patterns.
At the small scale (3B vs.\ 4B), both models obtain comparable deltas on Chart Data Parsing (+.032 vs.\ +.033) and the ChartQA benchmark (+.024 vs.\ +.037), while Qwen3-VL-4B benefits more on ChartQA-Pro (+.033 vs.\ +.024) and CharXiv (+.016 vs.\ +.009).
At the large scale (7B vs.\ 8B), Qwen2.5-VL-7B gains more on NL2Chart (+.036 vs.\ +.012) and Chart Data Parsing (+.030 vs.\ +.005), whereas Qwen3-VL-8B shows substantially larger improvements on the ChartQA benchmark (+.098 vs.\ +.046), ChartQA-Pro (+.052 vs.\ +.039), and CharXiv (+.075 vs.\ +.058).
Notably, Qwen3-VL-8B starts from a lower ChartQA baseline than Qwen2.5-VL-7B (.725 vs.\ .800), yet consistency training closes much of this gap (+.098), suggesting that the cycle-consistent objective is especially effective at compensating for weaker chart-understanding priors.

\textit{Cross-architecture generalization.}
LLaVA-OneVision-4B, which uses a different vision encoder and LLM backbone, also benefits consistently, with its largest gain on CharXiv (+.018).
The smaller absolute deltas compared to Qwen-family models are expected: LLaVA-OneVision already starts from a strong baseline (e.g., .842 on the ChartQA benchmark, the highest among all backbones before consistency training), leaving less headroom for improvement, a pattern consistent with the ceiling effect observed in \autoref{sec:abl_sft}.
Nonetheless, the consistent positive direction across all five benchmarks confirms that the consistency objective is architecture-agnostic and does not rely on Qwen-specific inductive biases.

\textit{CharXiv as a stress test.}
CharXiv contains charts from diverse scientific literature with visualization styles substantially different from the Vega-Lite charts used during training. Even so, all five backbones improve, with Qwen3-VL-8B reaching .472 (+.075), highlighting the strong out-of-distribution generalization of consistency-based learning.

\subsection{Training Steps Ablation}
\begin{figure}
    \centering
    \includegraphics[width=\linewidth]{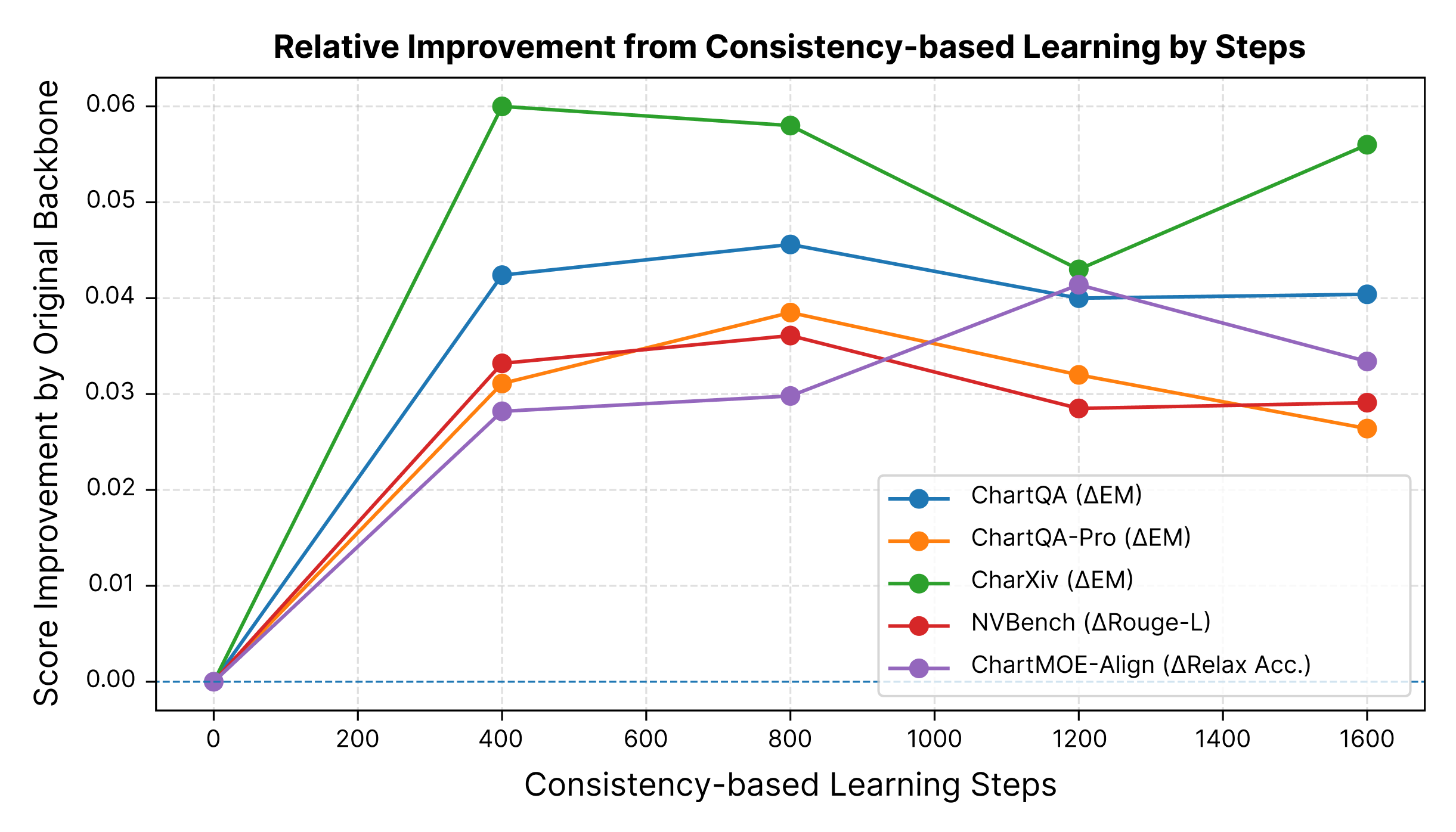}
    \caption{\textbf{Impact of consistency training steps on external benchmarks.} Performance rises steeply within the first 400--800 steps and then plateaus, indicating that the generate--parse consistency objective is highly sample-efficient.}
    \label{fig:steps}
\end{figure}

We analyze how training steps of consistency learning influence performance across tasks.
As shown in \autoref{fig:steps}, most benchmarks exhibit a steep improvement within the first 400--800 steps, after which the gains plateau or oscillate slightly. CharXiv shows the largest early boost, while the ChartQA benchmark, ChartQA-Pro, nvBench, and ChartMOE-Align all reach stable improvements in roughly the same range. This pattern indicates that CycleChart benefits quickly from consistency supervision: a relatively small number of steps is sufficient to align generation, parsing, and reasoning behaviors. Overall, the method achieves strong multi-task gains with modest training cost, demonstrating the efficiency of consistency-based learning.
This rapid convergence also makes early stopping practically important.
Because CycleChart trains predominantly on Vega-Lite specifications, prolonged training risks overfitting to Vega-Lite-specific syntax and layout conventions.
Early stopping at the plateau (400--800 steps) preserves the backbone's generalization ability, enabling the model to transfer to alternative output formats such as VQL, as demonstrated by the nvBench results in \autoref{tab:abl_xbench_backbones}.


\subsection{Cycle-Consistency vs.\ Multi-Task Fine-Tuning}
\label{sec:abl_sft}

A natural question is whether the gains of CycleChart stem from its cycle-consistency structure or simply from multi-task training on the same data.
To isolate this factor, we compare CycleChart-7B with a \textbf{Multi-Task SFT} baseline that is trained on the \emph{same} data and tasks but draws samples independently across tasks, breaking the per-sample alignment that underpins our generate--parse cycle.
Concretely, Multi-Task SFT shuffles the four task pools (NL2Chart, Schema Parsing, Data Parsing, ChartQA) so that each mini-batch may contain schema parsing labels from one chart and QA labels from another, whereas CycleChart always draws all four tasks from the same sample.

\begin{table}[t]
\centering
\small
\caption{Cycle-consistency vs.\ multi-task SFT on CycleChart-Bench.
Both models use identical data, tasks, and training budget; only the per-sample alignment differs.}
\label{tab:abl_sft}
\begin{tabularx}{\columnwidth}{>{\raggedright\arraybackslash}X ccc}
\toprule
\textbf{Task (Metric)} & \textbf{SFT-7B} & \textbf{CycleChart-7B} & $\boldsymbol{\Delta}$ \\
\midrule
NL2Chart (ROUGE-L \%)      & 84.75 & \textbf{87.12} & +2.37 \\
Schema Parsing (ROUGE-L \%) & 96.82 & \textbf{97.32} & +0.50 \\
Data Parsing (RNSS)         & 94.91 & \textbf{95.29} & +0.38 \\
ChartQA (EM)                & 72.09 & \textbf{75.50} & +3.41 \\
\bottomrule
\end{tabularx}
\end{table}

As shown in \autoref{tab:abl_sft}, CycleChart-7B consistently outperforms Multi-Task SFT across all four tasks.
The largest gain appears on the ChartQA task (+3.41 EM), suggesting that per-sample alignment between chart generation and parsing provides especially strong supervision for downstream reasoning.
NL2Chart also benefits substantially (+2.37 ROUGE-L), indicating that parsing feedback improves generation quality.
The more modest gains on Schema Parsing (+0.50) and Data Parsing (+0.38) are expected: both SFT and CycleChart already exceed 94 on these tasks, leaving little headroom; our data volume ablation (\autoref{tab:abl_data_volume}) similarly shows that structural tasks saturate early, confirming a ceiling effect rather than a weak signal.
Importantly, the benefit of per-instance alignment is amplified on out-of-distribution data: as shown in \autoref{tab:abl_xbench_backbones}, the same consistency objective yields up to +5.8\% EM on CharXiv and +4.6\% on the ChartQA benchmark, whose chart styles are entirely absent from training, whereas Multi-Task SFT, lacking per-instance alignment, cannot provide this form of structural generalization.
These results confirm that CycleChart's advantage is not merely due to multi-task training, but specifically due to the cycle-consistent alignment that couples generation and parsing within each instance.

\subsection{Data Volume Scaling}
\label{sec:abl_data_volume}

To understand how performance scales with dataset size, we train CycleChart-7B on 25\%, 50\%, 75\%, and 100\% of the CycleChart-Bench training set.
Unlike the main results (\autoref{tab:CycleChart_nl2chart}, \autoref{tab:CycleChart_schema}), which use a fixed budget of 2{,}000 steps, each variant here is trained to convergence to isolate the effect of data volume.
The 100\% converged model therefore slightly outperforms the fixed-step main model; we retain the fixed-step configuration as our primary result because it provides a fairer comparison across ablations.

\begin{table}[t]
\centering
\small
\caption{Performance at different data proportions on CycleChart-Bench.
Each model is trained to convergence.}
\label{tab:abl_data_volume}
\begin{tabularx}{\columnwidth}{>{\raggedright\arraybackslash}X cccc}
\toprule
\textbf{Task (Metric)} & \textbf{25\%} & \textbf{50\%} & \textbf{75\%} & \textbf{100\%} \\
\midrule
NL2Chart (ROUGE-L \%)        & 87.34 & 89.54 & 91.18 & \textbf{92.12} \\
Schema Parsing (ROUGE-L \%)  & 97.48 & 97.44 & 98.26 & \textbf{98.34} \\
Data Parsing (RNSS)          & 95.42 & 96.61 & 96.57 & \textbf{96.82} \\
ChartQA (EM \%)              & 77.20 & 80.62 & 83.56 & \textbf{85.11} \\
\bottomrule
\end{tabularx}
\end{table}

\autoref{tab:abl_data_volume} reveals two distinct scaling patterns.
Structural tasks (Schema Parsing and Data Parsing) reach near-saturation performance at just 25\% of the data, with gains of less than 1 point from 25\% to 100\%.
This high data efficiency suggests that the cycle-consistency objective provides strong inductive bias for recovering chart structure, requiring relatively few examples to learn the underlying Vega-Lite grammar.
In contrast, the ChartQA task exhibits steady improvement across all data proportions, rising from 77.20 to 85.11 EM, consistent with the greater diversity of reasoning patterns required for question answering.
NL2Chart follows an intermediate trend, improving consistently but with diminishing marginal returns.
These results indicate that consistency-based learning is inherently sample-efficient: CycleChart-Bench's 6,507 charts, which are orders of magnitude smaller than existing chart corpora such as ChartInstruct (191K) or UniChart (611K), are sufficient for structural tasks, and further data augmentation would primarily benefit reasoning-oriented tasks like ChartQA.

\subsection{Single-View vs.\ Faceted Chart Analysis}
\label{sec:abl_faceted}

CycleChart-Bench includes both single-view charts (5,372) and faceted multi-view charts (1,135).
To assess whether cycle-consistency training benefits more complex layouts disproportionately, we report a breakdown of CycleChart-7B's performance on these two subsets, comparing against the Qwen2.5-VL-7B-Instruct backbone.

\begin{table}[t]
\centering
\small
\caption{Single-view vs.\ faceted chart performance on CycleChart-Bench.
``Base'' denotes the Qwen2.5-VL-7B-Instruct backbone; ``Ours'' denotes CycleChart-7B.}
\label{tab:abl_faceted}
\setlength{\tabcolsep}{3pt}
\begin{tabularx}{\columnwidth}{>{\raggedright\arraybackslash}X >{\centering\arraybackslash}p{1.6cm} >{\centering\arraybackslash}p{0.7cm} >{\centering\arraybackslash}p{1.6cm} >{\centering\arraybackslash}p{0.7cm}}
\toprule
& \multicolumn{2}{c}{\textbf{Single-View}} & \multicolumn{2}{c}{\textbf{Faceted}} \\
\cmidrule(lr){2-3} \cmidrule(lr){4-5}
\textbf{Task (Metric)} & \textbf{Base$\to$Ours} & \textbf{Gain} & \textbf{Base$\to$Ours} & \textbf{Gain} \\
\midrule
NL2Chart (Valid \%)        & 40.9$\to$98.9 & +58.0 & 19.3$\to$99.2 & \textbf{+79.9} \\
Schema (ROUGE-L \%)       & 70.7$\to$97.5 & +26.8 & 64.3$\to$96.4 & \textbf{+32.1} \\
Data Parsing (RNSS)       & 68.5$\to$95.5 & +27.0 & 66.0$\to$94.1 & \textbf{+28.0} \\
ChartQA (EM \%)           & 58.3$\to$75.7 & \textbf{+17.4} & 59.1$\to$74.5 & +15.5 \\
\bottomrule
\end{tabularx}
\end{table}

\autoref{tab:abl_faceted} shows that CycleChart delivers larger absolute gains on the harder faceted subset for three out of four tasks.
The most striking difference is in NL2Chart validity, where the baseline achieves only 19.3\% on faceted charts (vs.\ 40.9\% on single-view), yet CycleChart raises both to near-perfect validity (99.2\% and 98.9\%).
Schema Parsing and Data Parsing show a similar pattern: the baseline struggles more with faceted charts, but CycleChart closes the gap, with gains of +32.1 and +28.0 on faceted charts compared to +26.8 and +27.0 on single-view.
The ChartQA task is the exception, where single-view gains (+17.4) slightly exceed faceted gains (+15.5), likely because single-view questions can directly benefit from improved generation and parsing, whereas faceted QA requires additional cross-panel reasoning that is not fully captured by the cycle objective alone.
Overall, these results confirm that cycle-consistency is especially effective for compositionally complex charts, where the structural priors enforced by the generate--parse loop most directly reduce specification errors.

\subsection{Qualitative Analysis}

\begin{figure*}[t]
\centering
\includegraphics[width=\textwidth]{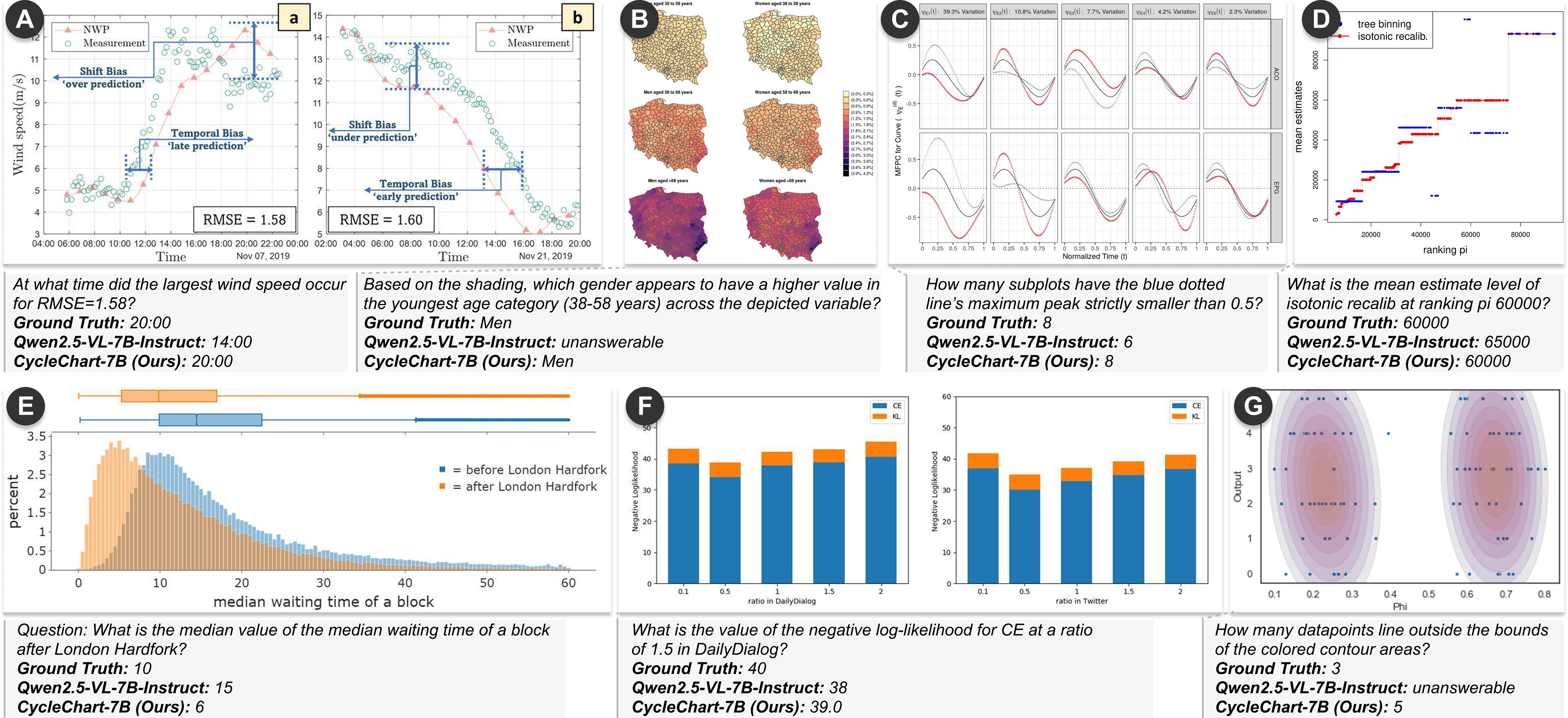}
\caption{Qualitative ChartQA cases from CharXiv~\cite{wang2024charxiv}, whose chart styles are absent from our training data.
\textbf{Top row (A--D):} CycleChart-7B corrects baseline errors, demonstrating improved numerical grounding~(A), categorical comparison in unseen chart types~(B), counting across faceted subplots~(C), and fine-grained value reading~(D).
\textbf{Bottom row (E--G):} Both models fail, revealing remaining challenges in glyph disambiguation for overlapping statistical marks~(E), numerical drift on dense axes~(F), and geometric region--point inclusion reasoning~(G).}
\label{fig:qual_cases}
\end{figure*}

\autoref{fig:qual_cases} presents representative ChartQA cases from CharXiv~\cite{wang2024charxiv}. Its figures, which include dense multi-panel layouts, heavily annotated scientific figures, and map-like demographic visualizations, are considerably more complex than those in our training corpus.
While our training data primarily consists of simple line, bar charts, and faceted layouts rendered via Vega-Lite, CharXiv includes chart types entirely absent during training, making it a strong test of generalization.

\paragraph{Successful reasoning (A--D).}
Case~A requires consistent axis alignment and careful comparison between visually similar curves across paired subplots.
The baseline misreads the temporal alignment and reports 14:00, whereas CycleChart retrieves the correct time 20:00.
This improvement likely stems from the schema parsing objective, which trains the model to attend to axis labels and tick positions as part of recovering the chart specification, thereby reinforcing temporal grounding.
Case~B involves a demographic visualization with subtle color variations, a chart type not seen during training.
The baseline reports it as unanswerable, suggesting that it fails to recognize the visual encoding entirely.
CycleChart correctly infers the gender category, demonstrating that the structural priors learned from Vega-Lite-based charts (e.g., color-to-category mappings) transfer to unseen visualization styles.
Case~C asks how many subplots satisfy a condition on the maximum of a dotted line.
The baseline undercounts (6 vs.\ ground truth 8), likely due to inconsistent attention across panels.
CycleChart correctly identifies all qualifying subplots, suggesting that training on faceted chart generation and parsing strengthens cross-panel awareness.
Case~D involves reading an isotonic calibration curve at a specific ranking position.
The baseline produces a plausible but incorrect estimate (65000 vs.\ ground truth 60000), whereas CycleChart returns the correct value.
This reflects that generate--parse consistency encourages precise visual grounding over heuristic interpolation, as the model must learn to recover exact data values during the data parsing task.

\paragraph{Remaining failure modes (E--G).}
Case~E involves a boxplot where the median line visually resembles axis tick marks.
Both models produce incorrect estimates (Base: 15, Ours: 6, GT: 10), with errors in opposite directions, suggesting that neither has learned a reliable representation of boxplot-specific glyphs.
Since our training data does not contain boxplots and the consistency objectives do not include glyph-type-specific structural cues, the model lacks the inductive bias needed to disambiguate the median line from other horizontal guides.
Case~F shows both models exhibiting moderate deviation when estimating a continuous value from a dense chart (Base: 38, Ours: 39.0, GT: 40).
CycleChart produces a closer estimate, indicating partial improvement in axis--mark alignment, but the residual error reveals a fundamental precision ceiling: when axis ticks are sparse relative to the data range, sub-tick interpolation remains unreliable for current multimodal LLMs.
Case~G requires determining whether scatter points lie outside contour regions, a region--point inclusion task that is fundamentally geometric rather than encoding-level.
The baseline reports it as unanswerable, while CycleChart attempts an answer (5) but overcounts (GT: 3), suggesting that it has learned to engage with spatial queries but lacks the geometric reasoning precision to resolve boundary cases.
This class of spatial reasoning, which involves containment, intersection, and proximity, lies outside the scope of our current generate--parse objectives and represents a complementary challenge for future chart understanding research.

\section{Discussion}

Our results demonstrate that generate--parse consistency is a powerful and general training principle for chart understanding.
Below we discuss the current limitations and promising extensions along four axes: data efficiency, grammar coverage, evaluation methodology, and offline versus online consistency.

\paragraph{Data efficiency.}
CycleChart-Bench contains only 6,507 charts, which is orders of magnitude smaller than ChartInstruct (191K) or UniChart (611K), yet CycleChart consistently outperforms models trained on those larger corpora.
This data efficiency is a direct consequence of the generate--parse consistency objective: because every sample simultaneously supervises forward generation and reverse parsing through a shared instance, each chart provides far denser supervision than an independently sampled multi-task example.
Our data scaling ablation (\autoref{tab:abl_data_volume}) further quantifies this effect: structural tasks saturate at just 25\% of the data, whereas the ChartQA task improves steadily from 77.20 to 85.11 EM across the full range, suggesting that consistency-based learning is inherently sample-efficient for structural tasks and that reasoning-oriented tasks benefit most from additional data.
A promising direction is to incorporate web-scraped or real-world charts with pseudo-specifications, which are generated by a strong VLM and verified through the cycle-consistency loop itself, to increase both visual and semantic diversity without requiring manual annotation.
Furthermore, the current QA set is constructed from a fixed set of question templates; augmenting it with open-ended, compositional questions (e.g., multi-hop comparisons across chart panels) would better stress-test the reasoning capabilities of future models.

\paragraph{Visualization grammar coverage.}
The benchmark is restricted to Vega-Lite-style charts, creating a visual distribution gap relative to charts rendered by other libraries (e.g., matplotlib, D3, or hand-drawn).
Our CharXiv transfer results (\autoref{tab:abl_xbench_backbones}) show that the structural priors learned through cycle-consistency partially bridge this gap, as CycleChart-7B achieves +5.8\% EM on CharXiv despite never seeing its chart styles. But generalization degrades for highly stylized or non-standard chart types such as infographics, 3D visualizations, and charts with heavy textual annotations.
Crucially, the generate--parse consistency framework is grammar-agnostic: the cycle objective only requires a deterministic renderer $R$ and a data extraction interface, both of which can be instantiated for any declarative grammar.
Extending CycleChart to ECharts or matplotlib would therefore involve swapping the rendering and extraction components while retaining the same training loop, enabling a single model to develop consistency priors across multiple visualization ecosystems.


\paragraph{Automated visual evaluation.}
Our current evaluation relies on reference-based metrics (PSNR, CLIP, MS-SSIM) that require ground-truth renderings.
While these metrics capture pixel-level fidelity and perceptual similarity, they are insensitive to semantically important but visually subtle differences. For example, a swapped legend color mapping produces a high PSNR but conveys entirely wrong information.
An LLM-as-judge paradigm, where a strong VLM evaluates whether a rendered chart faithfully conveys the intended data and structure, could complement reference-based metrics with semantic correctness judgments.
Such a judge could also serve as a reward signal for reinforcement-based fine-tuning, enabling self-improving consistency learning without pixel-level ground truth.
This is particularly relevant for scaling to grammars where deterministic rendering is unavailable or where multiple valid visual encodings exist for the same data.

\paragraph{Offline versus online consistency.}
CycleChart enforces consistency offline: reverse-path targets are pre-computed from gold specifications rather than from the model's own predictions.
This is deliberate---rendering every predicted $\hat{s}$ during training would introduce substantial I/O overhead and frequent failures from invalid specifications, especially early in training when validity is low (\autoref{tab:abl_faceted}).
The offline design produces noise-free targets, and per-instance alignment still ensures that forward and reverse tasks share the same data instance, which is the key property distinguishing CycleChart from standard multi-task training (\autoref{tab:abl_sft}).
An online or hybrid schedule---switching to on-the-fly rendering once validity stabilizes---could tighten the cycle further and is a promising direction for future work.


\section{Conclusion}

We presented CycleChart, built on the observation that chart generation and understanding are two sides of the same coin: forcing a model to close the generate--parse loop on every instance yields far stronger representations than multi-task training on the same tasks independently.
Our experiments surface three insights:
\emph{consistency over multi-tasking}---per-instance alignment, not task diversity, drives the gains;
\emph{renderer as free annotator}---deterministic Vega-Lite execution derives all reverse-path targets at zero labeling cost;
and \emph{compositionality as the bottleneck}---the cycle objective benefits compositionally complex faceted charts most, indicating that it teaches structure, not surface patterns.
Together with strong transfer to unseen external benchmarks, these results position CycleChart and CycleChart-Bench as solid foundations for future chart understanding research.


\bibliographystyle{abbrv-doi-hyperref}

\bibliography{main}

\end{document}